# Emoji Sentiment Scores of Writers using Odds Ratio and Fisher Exact Test

JOSE BERENGUERES, UAE University

The sentiment of a given emoji is traditionally calculated by averaging the ratings {-1, 0 or +1} given by various users to a given context where the emoji appears. However, using such formula complicates the statistical significance analysis particularly for low sample sizes. Here, we provide sentiment scores using odds and a sentiment mapping to a 4-icon scale. We show how odds ratio statistics leads to simpler sentiment analysis. Finally, we provide a list of sentiment scores with the often-missing exact p-values and CI for the most common emoji.

## 1 INTRODUCTION

Correctly estimating sentiment of a text can enhance the effectiveness of applications that range from automated psychological support such as the woeBot chatbot, to suicide and depression prevention systems [1, 2]. In addition, sentiment scores are useful in user profiling and recommender systems. However, a bias in the score can introduce bias in the outcome, and as sentiment use becomes more pervasive it can potentially impact the lives of many people. Therefore, it is important from an engineering and ethical perspective to understand such biases. To give a few examples, [3] correlated the emoji people use with the economic development. Emoji can also be used to profile the gender of the author of a tweet [4] and, more recently, emoji usage has been correlated with higher levels of engagement and happiness at work [5]. In the same vein, Tweets that contain emoji score higher sentiment than tweets without emoji [6].

### 1.1 How sentiment is estimated

One (out of the many) ways to estimate the sentiment of a text, is to count the number of negative and positive items (words or emoji) and then compute average. To do this, first we need to know the sentiment score of each word or emoji that are available in public benchmarks. Such benchmarks are usually built by asking humans to read and rate texts from -1 to +1 or by assigning values based on linguistic assumptions (AFINN, etc.). A popular emoji benchmark was compiled by [6]. Machine learning models have also been applied to build Emoji benchmarks [7]. The differences of scores between the writer and the reader have also been investigated [5].

### 1.2 Sources of bias

Various authors have identified factors that can affect the accuracy and precision of the sentiment score of a word or emoji. For example, different phone makers and operating systems will display the same Emoji code differently. Apple's emoji images are different from the original NTTdocomo Emoji set, this impacts how emoji are perceived. [8] studied the variations in interpretation depending on the Emoji set used (Apple, Android…). Finally, other potential sources of bias are found in how the scores themselves are calculated.

## 1.3 Score formula

As we mentioned, to compute the overall sentiment score of a text, individual sentiment scores of the keywords appearing in the text are aggregated into a single number that best estimates the sentiment. Usually the average function is used, but other linear and non-linear estimators exist (median, max, machine learning models). These keyword-sentiment pairs can be found in tables that are pre-computed by showing texts to users and then eliciting a rating about the text that contains the keyword. To build a reliable table, this process is repeated with N individuals and M texts, often using Amazon Mechanical Turk. The sentiment of a keyword is the number that results from calculating the mean of individual ratings when grouped by keyword. In other words, the sentiment of a keyword can be interpreted as the expected sentiment of the ***texts*** where the keyword appears. Thus, if a keyword such as *wonderful* appears in $M$ texts and we show these $M$ texts to a $N$ people, then the sentiment can be calculated as

$$S_{wonderful} = \frac{1}{NM} \sum_{i=0}^{N} \sum_{j=0}^{M} s_{ij} \tag{1}$$

where $s_{ij}$ is the rating by $i^{th}$-user of the $j^{th}$-text where the keyword *wonderful* appeared and text is typically a Tweet or comment. Usually, sentiment is rated in a unidimensional bounded sentiment scale [+1, -1], it can be discreet or continuous. This has the benefit of simplicity but also has drawbacks. For example, given an ambivalent keyword such as *crazy* that appears in positive and negative texts, its score is close to zero even though, the word is mostly used in non-neutral contexts.

## 1.4 Low sentiment

This bias to zero effect is more pronounced in emoji, as emoji tend to be used more often than words as modifiers, disambiguators and qualifiers of speech [9]; as in *very* good or *very* bad. A typical example is, adding a clapping hands emoji or flamenco dance to a text to emphasize the intensity of the text. Other ways emoji are used differently from words is to add a smile after a written request to increase the agreeableness [10] of the requester and to defuse potential tension. Another example of usage for emphasis is the emoji 631 − scream face - 😱. This emoji has an empirical s score of +0.200 and +0.190 by [5, 6]. However, the sentiment standard distribution is 0.791, one of the highest in the benchmark. This is explained by the fact that is used in both positive and negative contexts. In contrast, emoji 499 − blue heart - 💙 has empirical score of +0.864 and 0.730 by [5,6] with a much lower 0.40 standard deviation. The averaging to zero phenomena can be observed in many emoji that have sentiment scores close to zero but have standard deviation of sentiment higher than the average. Other similar cases are: 🔥, 😭, 😬, 👎, and 😔. These emoji have relatively high SD and low absolute $S$ sentiment. Given this, emoji based sentiment analysis can increase accuracy if more granularity was considered. The current averaging approach to compute sentiment scores seems to oversimplify nuances, and the high number of neutral emoji seems to support this claim [5, 6, 8]. To illustrate this point, of Fig. 1 shows four histograms of emoji. X-axis is sentiment rating using the 4-emoji scale, 1 is sad, 4 is great. Note that LOL cry face emoji has a low absolute $S$ sentiment due to the flatter histogram.

## 1.5 Odds ratio

Following, we show a way to compute sentiment scores that does not suffer from the averaging to zero effect of Eq.1. Ratings for the positive sentiment dimension, the negative sentiment dimension as well as the neutral sentiment dimension are separated. For each dimension a mean is not provided, instead and odds ratio together with a Fisher test p-value is provided. Following we show examples of how using odds ratio instead $S$ scores can lead to more accurate overall sentiment assessment of content.

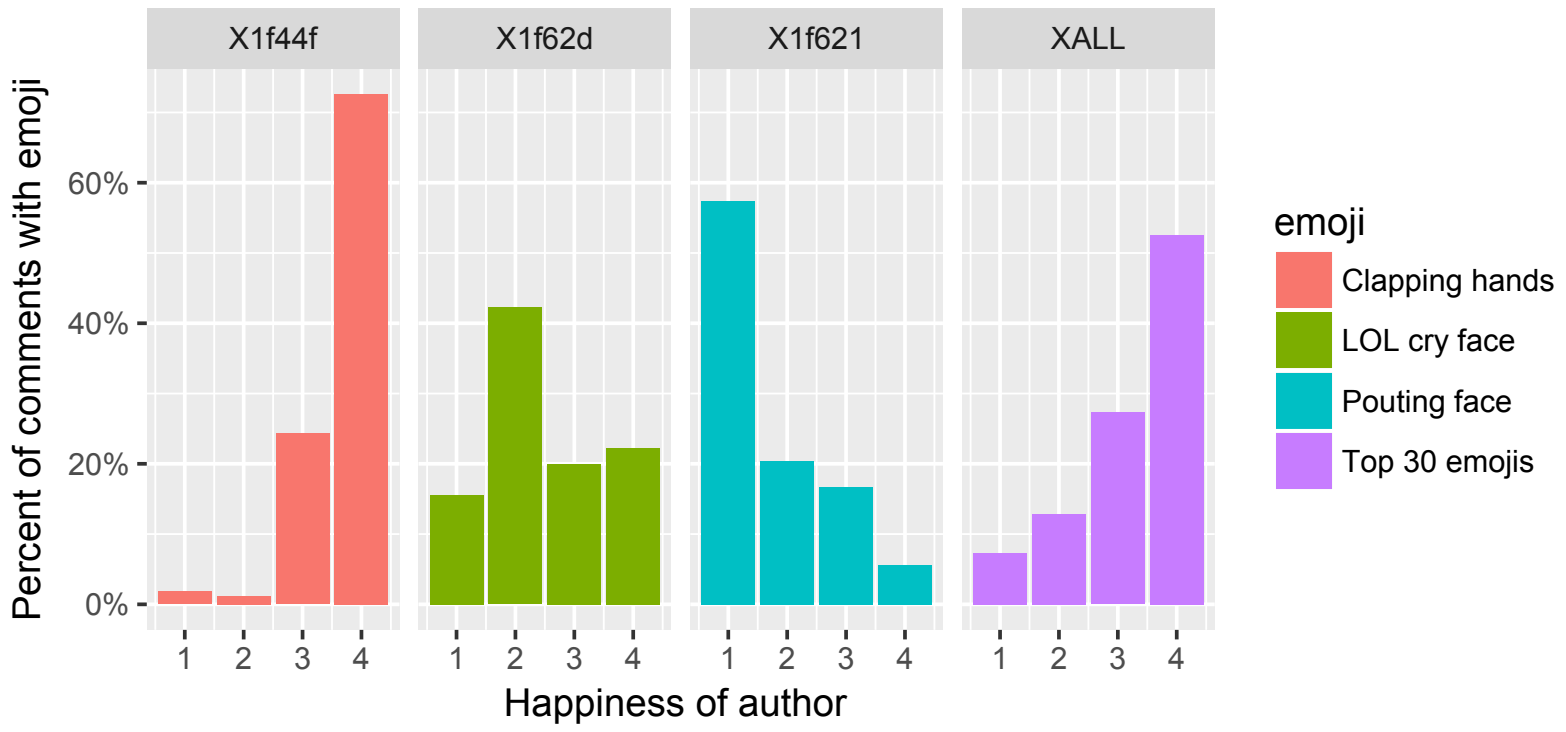

**Fig. 1. The S sentiment score fails to capture nuances in the distribution of sentiment of certain emoji such as the LOL cry face that do not follow a normal distribution. Reprinted from [5]. Label, 1: Sad, 2: Neutral, 3: Good, 4: Great**

# 2 EXPERIMENTAL AND COMPUTATIONAL DETAILS

## 2.1 Data set description

Here, we use the same data set as in [11], but with new samples that have been added since the time of publication. The dataset and a R notebook kernel can be found at https://www.kaggle.com/harriken/odds-ratio-sentiment-emoji

## 2.2 Data collection

Data used here was collected from 2014-05-10 to 2017-03- 08. The data was generated by employees that work(ed) at 56 different companies. The companies belong to one of the following sectors: e-payment start-up, IT consulting services, retail, manufacturing, services, tourism or education. About half of the companies are multinational and the other half are Barcelona-based companies. Employee data was collected in the framework of corporate feedback as part of their companies' kaizen initiatives. The bulk of the employees used this mobile application in Spain (Barcelona area). More than 90% are Spanish nationals. The comments are written in various languages: 97% in Spanish, 2% in English, 1% Catalan. The data consisted of two tables: vote-sentiment and comments with emoji, likes, employee metadata. Here, we will focus on two of them: vote-sentiment and comments. An in-depth description of the dataset is available at [5,10].

## 2.3 User flow

A user ***sentiment*** score was obtained when an employee opened the app and answered the question: How happy are you at work today? To answer, the employee indicates their feeling by touching one of four icons that appear on the screen. See Table 1. The UI of the English version is shown in the table. The default UI was in Spanish language. After the employee indicates their happiness level, a second screen appears where they can input a text explanation (usually a suggestion or comment), this is the comments table. Finally, in a third screen employees can see their peers' comments and like or dislike them. In total, 68k comments were recorded with a median length of 58 chars per comment. 5% of comments contain emoji. 962 users used emoji at least once, 4,893 users never used emoji. Out of 63k comments, 3,506 comments contain at least one emoji. 358 different types of emoji are used which appeared 10,048 times. There are 2,666 unique emoji in the Unicode Standard as of June 2017).

## 2.4 Emoji to 4-emoji as sentiment mapping

*2.2.1 Emoji to numeric sentiment mapping.* [5] provided sentiment scores from the writer point of view using the formula of Eq.1 by mapping four categorical sentiment states to a numeric range {-1, -0.5, 0, +1} corresponding to the emoji-icons sad, neutral, good, and great. The effect of quantization noise was also discussed. The discretization values where chosen arbitrarily, this is also a potential source of bias.

*2.2.2 Emoji to emoji sentiment mapping.* Here, we map the sentiment of the user to a 4–icon sentiment scale via the 4-icon set as described in Table 1. For convenience, we assign the following text labels to each of the four Emoji of table 1 as *Sad, for the sad face, Neutral (for the neutral face icon), Good (smile face icon), Great (big smile face icon).* However, this labeling does not introduce any bias as the numeric mapping might do. Note that colors and position of the icons might be introducing some UX bias that is not assessed.

**Table 1. Data collection flow**

| Table **(Rows)** | **Feed-back** | **UI flow** |
|---|---|---|
| Happiness votes (398k) | **How happy are you at work today?**<br>**-** Great: yellow<br>- Good: green<br>- Neutral: purple<br>- Sad: red | ↓ 1st screen<br>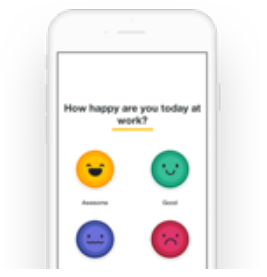 |
| Comments (68,476) | **Comment box** | ↓ 2nd screen<br>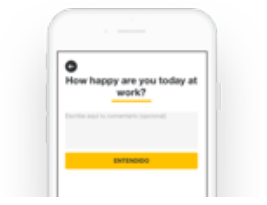 |
| Likes / Dislikes (599,103) | **Anonymous forum**<br>Users can:<br>- **view** comments<br>- **like** a comment<br>- **dislike** a comment | ↓ 3rd screen<br>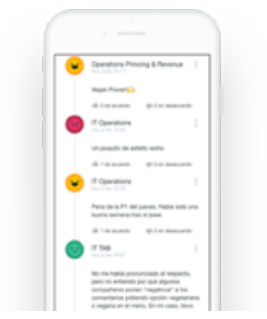 |

*Source:* myHappyforce.com

# 3 RESULTS AND DISCUSSION

## 3.1 Odds sentiment via 4-icon mapping

*3.1.1 Odds ratio.* Given an emoji, the odds ratio is defined as the odds that the user selected a given sentiment-icon divided by the ground truth odds. In this case, we take the ground truth as the distribution of sentiment icons of the dataset. We have a total sample of 3680 comments with emoji, rated by its author with one or more of the 4-sentiment icons**, (Great:1870 counts, Good: 1017, Neutral: 505, Sad: 288)**. We observe a bias towards *Grea*t, which is explained in detail in [5] (engagement bias). Hence, for each emoji, we can compute the odds to elicit one of the four responses. The ratio compares a given odd to the ground truth odds. The ground truth can also be interpreted as the a priori odds.

*3.1.2 Example.* For example, emoji 🙂, was used in 33 occasions of which the sentiment mapping is as follows (Great:18 counts, Good: 11 counts, Neutral: 2, Sad: 2). The odds given 🙂 that the sentiment is ***Great*** is ***1.2*** *= 18/(33-18).* The ground truth of odds for *Great* is *1.033 = 1870 / (3680-1870);* Hence, the odds **ratio** for *Great* given 🙂 is approximately 1.2. The odds ratio that the sentiment is *Good* is 2.1, the odds ratio for *Neutral* is 0.4 and the odds ratio for *Sad* is 0.76. Applying a Fisher exact test for count data [12], we can see that none of these odds is significant with a p-value lower than 0.05 except for the odds ratio for the *Good* sentiment at 2.1, meaning that that the odds that the sentiment will be *Good* relative to the a priori odds for Good is 2.1 times. The a priori odds of Good is 0.38. Converting from odds to probability is straightforward using the odds definition *a/(1-a).* Table 2 compares two ways to compute sentiment scores: 1) the numerical method used in [5,6], and Eq.1, and 2) the odds ratio sentiment via 4-icon mapping. Table 3 lists the odds with the corresponding Fisher test results. Of all the cases, only 78 emoji with odds with p-values$<0.05$, are listed.

## 3.2 Comparison of numerical sentiment vs. odds

From Table 2 we can see how both scoring systems agree in general. The *S* score listed is calculated with Eq.1. Ratios are omitted when the p-score was less than 0.05. The scores are calculated using the data from Table 3.

*3.2.1 Fire emoji.* For example, row 10 in Table 2 lists data for the 🔥 emoji, this emoji has an *S* =+0.09, a score close to zero. However, even though in average the sentiment is positive, according to the Fisher test this emoji has no predictive value as a positive sentiment estimator (Great, Neutral, or Good). However, it does have predictive value as an estimator for negative sentiment (Sad), with a ratio of 3.9, CI (1.2, Inf) and p-value to reject the null hypothesis that both odds (compared to the a priori odds) are the same of 0.03.

*3.2.2 Borderline case emoji.* Another example is 🤷‍♀️, this emoji gets an *S* score of -0.67 (Table 3), quite negative considering the minimum is -1. This is due to the low sample number of 3 samples only. However, the Fisher test results estimates an odds ratio of 12 for the *Neutral* sentiment vs the a priori odds, the CI is (1,Inf). In this case the predictive power towards *neutral* sentiment is not statistically zero. While this row would be usually discarded as borderline case using other methods, the Fisher test shows its superior accuracy to asses significance. Finally let's examine odds as probability.

*3.2.3 Odds as probability.* 🙋, gets an *S* = +0.76, relatively high. However, how should we interpret this number? On the other hand, the odds ratio for *Great*-sentiment is 5.8 with a CI (2.1, Inf). This information is readily usable in form of probability. When🙋 appears in a comment the writer is 5.8 times more likely to rate his sentiment as Great than if a random emoji appeared. Note that this is not necessarily true for every emoji that scores S>0.76.

*3.2.4 Collocated emoji.* In both tables, we can observe sometimes pairs and triplets of emoji, these are considered as one case. If the same emoji is repeated more than once in a comment it counts as 1.

*3.2.5 Multiple Odds.* In general, the Fisher test is only significative for one of the 4 possible sentiments. However, some Emoji are significant in odds ratio in more than one sentiment case. Such is the case of 😫, that has significant odds for the sentiments of *Great*, *Good*, and *Sad* but not for *Good*. See 4$^{th}$ row in Table 2. In the case of *Sad, the ratio* is 6.9, 3.7 for *Neutral*, and 1/5$^{th}$ for the odds ratio of feel *Great*.

# 4 CONCLUSIONS

In summary, using a relatively small dataset of 10k emoji, we have provided an alternative way to compute sentiment scores using odds instead of the traditional averaging numeric '*S*' formula. The use of odds ratio has the advantage that the powerful Fisher Exact test for count data provides exact p-values and exact CI's, whereas when the *S* score is used, the p-values obtained rely on approximations that are not accurate or valid when certain assumptions about the data are not true (normality). We have also provided and validated a way to rate sentiment with graphical icons. We have compared the results and ranked emoji by sentiment using both methods (numeric S vs. icon mapping). Both methods agree on the polarity of the sentiment in more than 90% of cases. A third advantage of using odds sentiment vs. the *S* numeric score is that odds can be easily used as probabilities while this is not true for *S*. Finally, a fourth advantage of odds is that because a ratio can be calculated relative to a ground truth of odds (in our case we used the a priori sentiment distribution over all the corpus), the ratio does not suffer distribution biases. Hence, differential sentiment analysis inherently robust and straightforward even with low sample sizes. We hope this new way of describing, using and computing sentiment scores will help simplify sentiment analysis in a variety of contexts.

**Table 2. Comparison of *S* score to icon-mapped odds ratio sentiment**

| | S | | Odds ratio | | | | |
|---|---|---|---|---|---|---|---|
| Emoji | Mean | SD | sad | neutral | good | great | Hex |
| 🤢 | -0.83 | 0.29 | 23.5 | | | | 1f922 |
| □ | -0.38 | 0.95 | 11.8 | | | | 1f593 |
| 🖕 | -0.25 | 0.96 | 11.8 | | | | 1f595 |
| 😡 😤 | -0.75 | 0.29 | 11.8 | | | | 1f624 1f621 |
| 😫 | -0.45 | 0.62 | 6.9 | 3.7 | | 0.2 | 1f62b |
| 😤 | -0.57 | 0.39 | 6.5 | 4.7 | | | 1f624 |
| 😔 | -0.38 | 0.59 | 4.6 | | | 0.2 | 1f614 |
| 🙁 | -0.36 | 0.61 | 4.5 | 4 | | 0.2 | 1f641 |
| 😣 | -0.41 | 0.54 | 4.1 | 5 | | 0.1 | 1f623 |
| 🔥 | 0.09 | 0.88 | 3.9 | | | | 1f525 |
| 😥 | -0.32 | 0.63 | 3.7 | | | 0.3 | 1f625 |
| 😭 | -0.27 | 0.62 | 2.9 | | | 0.3 | 1f62d |
| 😢 | -0.25 | 0.68 | 2.8 | | 0.4 | 0.4 | 1f622 |
| 😱 | 0.18 | 0.82 | 2.6 | | | | 1f631 |
| 🤔 | -0.08 | 0.64 | 2.3 | 1.7 | 2 | 0.4 | 1f914 |
| 👍 | 0.65 | 0.54 | 0.4 | | | | 1f44d |
| 🙌 | 0.68 | 0.53 | 0.3 | | | | 1f64c |
| 😘 | 0.60 | 0.54 | 0.2 | | | | 1f618 |
| 🤦 | -0.42 | 0.2 | | 31.4 | | | 1f926 |
| 😔 😭 | -0.63 | 0.25 | | 18.8 | | | 1f62d 1f614 |
| 😮 | 0.00 | 0.87 | | 12.6 | | | 1f62e |
| 🙉 | 0.00 | 0.87 | | 12.6 | | | 1f649 |
| 😷 🤒 🤧 | -0.67 | 0.29 | | 12.6 | | | 1f927 1f912 |
| 🤷 | -0.67 | 0.29 | | 12.6 | | | 1f937 |
| 😟 | -0.39 | 0.6 | | 7.9 | | | 1f61f |
| 😷 | -0.33 | 0.54 | | 6.3 | | | 1f637 |
| 😴 | -0.17 | 0.67 | | 5.5 | | | 1f634 |
| 😨 | -0.23 | 0.68 | | 5.2 | | | 1f628 |
| 😳 | -0.09 | 0.71 | | 4.9 | | | 1f633 |
| 🙈 | 0.04 | 0.74 | | 4.5 | | | 1f648 |
| 😜 | 0.59 | 0.57 | | 0.2 | | | 1f61c |
| 🎈 🎂 🎁 | 0.00 | 0 | | | Inf | | 1f381 1f382 |
| 💥 🎂 | 0.20 | 0.45 | | | 10.5 | | 1f382 1f4a5 |
| 😗 | -0.20 | 0.45 | | | 10.5 | | 1f617 |
| 🎂 🎁 | 0.22 | 0.44 | | | 9.2 | | 1f381 1f382 |
| 🎂 | 0.29 | 0.47 | | | 6.3 | | 1f382 |
| 🍾 | 0.33 | 0.52 | | | 5.2 | | 1f37e |
| □ | 0.17 | 0.58 | | | 5.2 | | 1f592 |

| | | | | | | |
|---|---|---|---|---|---|---|
| 🙄 | -0.05 | 0.53 | 3.8 | | 0.3 | 1f644 |
| 🙏 | 0.18 | 0.64 | 2.2 | | | 1f64f |
| 😬 | 0.20 | 0.62 | 2.1 | | | 1f62c |
| 😑 | -0.35 | 0.56 | 0.3 | | 0.2 | 1f611 |
| 🌴 | 1.00 | 0 | | Inf | | 1f334 |
| 🍑 | 1.00 | 0 | | Inf | | 1f351 |
| 😘🎉🎂 | 1.00 | 0 | | Inf | | 1f382 1f389 |
| 🎈 | 1.00 | 0 | | Inf | | 1f388 |
| 💪🎉 | 1.00 | 0 | | Inf | | 1f389 1f4aa |
| 🏄 | 1.00 | 0 | | Inf | | 1f3c4 |
| 🏳 | 1.00 | 0 | | Inf | | 1f3f3 |
| 🐘 | 1.00 | 0 | | Inf | | 1f418 |
| 🐸 | 1.00 | 0 | | Inf | | 1f438 |
| 🐘👍 | 1.00 | 0 | | Inf | | 1f44d 1f418 |
| 🙌👍 | 1.00 | 0 | | Inf | | 1f44d 1f64c |
| 💃 | 1.00 | 0 | | Inf | | 1f483 |
| 💯 | 1.00 | 0 | | Inf | | 1f4af |
| 😹 | 1.00 | 0 | | Inf | | 1f639 |
| 🏊 | 0.83 | 0.41 | | | 8.9 | 1f3ca |
| 👆 | 0.83 | 0.41 | | | 8.9 | 1f446 |
| 💶 | 0.83 | 0.41 | | | 8.9 | 1f4b6 |
| 🤘 | 0.78 | 0.48 | | | 7.7 | 1f918 |
| 💙 | 0.80 | 0.42 | | | 7.1 | 1f499 |
| 💚 | 0.75 | 0.54 | | | 7.1 | 1f49a |
| 💡 | 0.80 | 0.41 | | | 7.1 | 1f4a1 |
| 😚 | 0.80 | 0.41 | | | 7.1 | 1f61a |
| 🙋 | 0.76 | 0.44 | | | 5.8 | 1f64b |
| 🍻 | 0.75 | 0.46 | | | 5.3 | 1f37b |
| 👊 | 0.73 | 0.47 | | | 4.7 | 1f44a |
| 💪👍 | 0.70 | 0.48 | | | 4.2 | 1f44d 1f4aa |
| 🙆 | 0.60 | 0.7 | | | 4.2 | 1f646 |
| 🎄 | 0.66 | 0.54 | | | 3.9 | 1f384 |
| 🙃 | 0.58 | 0.61 | | | 3.2 | 1f643 |
| 😝 | 0.56 | 0.63 | | | 3 | 1f61d |
| 🤓 | 0.62 | 0.5 | | | 2.9 | 1f913 |
| 🔝 | 0.58 | 0.57 | | | 2.8 | 1f51d |
| 🙂 | 0.45 | 0.65 | | | 2.1 | 1f642 |
| 😓 | -0.24 | 0.58 | | | 0.3 | 1f613 |
| 😖 | -0.28 | 0.55 | | | 0.2 | 1f616 |

**Table 3. Fisher Exact test results for count data with p-values**

| Odds type | Collocated Emoji | Great-icon | Good-icon | Neutral-icon | Sad-icon | Total count | Odds ratio (see column one) | CI 95% | p-value. Fisher exact test | Sentiment S | SD S | Hex |
|---|---|---|---|---|---|---|---|---|---|---|---|---|
| sad | 🤢 | 0 | 0 | 1 | 2 | 3 | 23.5 | (1.8,Inf) | 0.02 | -0.83 | 0.29 | 1f922 |
| sad | 🖕 | 1 | 1 | 0 | 2 | 4 | 11.8 | (1.3,Inf) | 0.03 | -0.25 | 0.96 | 1f595 |
| sad | | 1 | 0 | 1 | 2 | 4 | 11.8 | (1.3,Inf) | 0.03 | -0.38 | 0.95 | 1f593 |
| sad | 😡😤 | 0 | 0 | 2 | 2 | 4 | 11.8 | (1.3,Inf) | 0.03 | -0.75 | 0.29 | 1f624 1f621 |
| sad | 😫 | 2 | 3 | 7 | 7 | 19 | 6.9 | (2.7,Inf) | 0.0004 | -0.45 | 0.62 | 1f62b |
| sad | 😤 | 0 | 3 | 6 | 5 | 14 | 6.5 | (2.1,Inf) | 0.003 | -0.57 | 0.39 | 1f624 |
| sad | 😔 | 4 | 8 | 16 | 11 | 39 | 4.6 | (2.3,Inf) | 0.0002 | -0.38 | 0.59 | 1f614 |
| sad | 🙁 | 2 | 4 | 7 | 5 | 18 | 4.5 | (1.5,Inf) | 0.01 | -0.36 | 0.61 | 1f641 |
| sad | 😣 | 2 | 6 | 12 | 7 | 27 | 4.1 | (1.7,Inf) | 0.004 | -0.41 | 0.54 | 1f623 |
| sad | 🔥 | 7 | 2 | 3 | 4 | 16 | 3.9 | (1.2,Inf) | 0.03 | 0.09 | 0.88 | 1f525 |
| sad | 😥 | 8 | 11 | 25 | 14 | 58 | 3.7 | (2.1,Inf) | 0.0001 | -0.32 | 0.63 | 1f625 |
| sad | 😭 | 8 | 12 | 24 | 11 | 55 | 2.9 | (1.5,Inf) | 0.004 | -0.27 | 0.62 | 1f62d |
| sad | 😢 | 11 | 7 | 28 | 11 | 57 | 2.8 | (1.5,Inf) | 0.005 | -0.25 | 0.68 | 1f622 |
| sad | 😱 | 15 | 6 | 6 | 6 | 33 | 2.6 | (1.0,Inf) | 0.04 | 0.18 | 0.82 | 1f631 |
| sad | 🤔 | 15 | 34 | 17 | 13 | 79 | 2.3 | (1.3,Inf) | 0.009 | -0.08 | 0.64 | 1f914 |
| sad | 👍 | 13 | 57 | 2 | 6 | 20 | 0.4 | (0.0,0.7) | 0.004 | 0.65 | 0.54 | 1f44d |
| sad | 🙌 | 60 | 21 | 1 | 2 | 84 | 0.3 | (0.0,0.9) | 0.04 | 0.68 | 0.53 | 1f64c |
| sad | 😘 | 10 | 57 | 4 | 3 | 17 | 0.2 | (0.0,0.5) | 0.0006 | 0.60 | 0.54 | 1f618 |
| neutr | 🤦 | 0 | 1 | 5 | 0 | 6 | 31.4 | (4.5,Inf) | 0.0003 | -0.42 | 0.2 | 1f926 |
| neutr | 😔😭 | 0 | 0 | 3 | 1 | 4 | 18.8 | (2.1,Inf) | 0.009 | -0.63 | 0.25 | 1f62d 1f614 |
| neutr | 😷🤒🤧 | 0 | 0 | 2 | 1 | 3 | 12.6 | (1.0,Inf) | 0.05 | -0.67 | 0.29 | 1f927 1f912 1f637 |
| neutr | 🤷 | 0 | 0 | 2 | 1 | 3 | 12.6 | (1.0,Inf) | 0.05 | -0.67 | 0.29 | 1f937 |
| neutr | 😮 | 1 | 0 | 2 | 0 | 3 | 12.6 | (1.0,Inf) | 0.05 | 0.00 | 0.87 | 1f62e |
| neutr | 🙉 | 1 | 0 | 2 | 0 | 3 | 12.6 | (1.0,Inf) | 0.05 | 0.00 | 0.87 | 1f649 |
| neutr | 😟 | 1 | 1 | 5 | 2 | 9 | 7.9 | (2.1,Inf) | 0.004 | -0.39 | 0.6 | 1f61f |
| neutr | 😷 | 1 | 3 | 6 | 2 | 12 | 6.3 | (2.0,Inf) | 0.003 | -0.33 | 0.54 | 1f637 |
| neutr | 😴 | 3 | 3 | 7 | 2 | 15 | 5.5 | (2.0,Inf) | 0.002 | -0.17 | 0.67 | 1f634 |
| neutr | 😨 | 2 | 2 | 5 | 2 | 11 | 5.2 | (1.6,Inf) | 0.01 | -0.23 | 0.68 | 1f628 |
| neutr | 😣 | 2 | 6 | 12 | 7 | 27 | 5 | (2.4,Inf) | 0.0001 | -0.41 | 0.54 | 1f623 |
| neutr | 😳 | 4 | 3 | 7 | 2 | 16 | 4.9 | (1.8,Inf) | 0.003 | -0.09 | 0.71 | 1f633 |
| neutr | 😤 | 0 | 3 | 6 | 5 | 14 | 4.7 | (1.6,Inf) | 0.008 | -0.57 | 0.39 | 1f624 |
| neutr | 🙈 | 8 | 4 | 10 | 2 | 24 | 4.5 | (2.0,Inf) | 0.0008 | 0.04 | 0.74 | 1f648 |
| neutr | 🙁 | 2 | 4 | 7 | 5 | 18 | 4 | (1.6,Inf) | 0.007 | -0.36 | 0.61 | 1f641 |
| neutr | 😫 | 2 | 3 | 7 | 7 | 19 | 3.7 | (1.4,Inf) | 0.01 | -0.45 | 0.62 | 1f62b |
| neutr | 🤔 | 15 | 34 | 17 | 13 | 79 | 1.7 | (1.0,Inf) | 0.04 | -0.08 | 0.64 | 1f914 |
| neutr | 😜 | 42 | 20 | 2 | 2 | 66 | 0.2 | (0.0,0.6) | 0.004 | 0.59 | 0.57 | 1f61c |
| good | 🎈🎂🎁 | 0 | 3 | 0 | 0 | 3 | Inf | (1.5,Inf) | 0.02 | 0.00 | 0 | 1f381 1f382 1f388 |
| good | 😗 | 0 | 4 | 0 | 1 | 5 | 10.5 | (1.4,Inf) | 0.02 | -0.20 | 0.45 | 1f617 |
| good | 💥🎂 | 1 | 4 | 0 | 0 | 5 | 10.5 | (1.4,Inf) | 0.02 | 0.20 | 0.45 | 1f382 1f4a5 |
| good | 🎂🎁 | 2 | 7 | 0 | 0 | 9 | 9.2 | (2.1,Inf) | 0.003 | 0.22 | 0.44 | 1f381 1f382 |
| good | 🎂 | 5 | 12 | 0 | 0 | 17 | 6.3 | (2.4,Inf) | 0.0003 | 0.29 | 0.47 | 1f382 |
| good | | 3 | 8 | 0 | 1 | 12 | 5.2 | (1.7,Inf) | 0.006 | 0.17 | 0.58 | 1f592 |
| good | 🍾 | 2 | 4 | 0 | 0 | 6 | 5.2 | (1.0,Inf) | 0.05 | 0.33 | 0.52 | 1f37e |
| good | 🙄 | 3 | 13 | 4 | 2 | 22 | 3.8 | (1.7,Inf) | 0.002 | -0.05 | 0.53 | 1f644 |
| good | 🙏 | 15 | 21 | 7 | 3 | 46 | 2.2 | (1.3,Inf) | 0.007 | 0.18 | 0.64 | 1f64f |
| good | 😬 | 9 | 12 | 5 | 1 | 27 | 2.1 | (1.0,Inf) | 0.05 | 0.20 | 0.62 | 1f62c |
| good | 🤔 | 15 | 34 | 17 | 13 | 79 | 2 | (1.3,Inf) | 0.003 | -0.08 | 0.64 | 1f914 |
| good | 😢 | 11 | 7 | 28 | 11 | 57 | 0.4 | (0.0,0.7) | 0.005 | -0.25 | 0.68 | 1f622 |
| good | 😑 | 3 | 3 | 16 | 4 | 26 | 0.3 | (0.0,1.0) | 0.05 | -0.35 | 0.56 | 1f611 |
| great | 🐸 | 7 | 0 | 0 | 0 | 7 | Inf | (3.3,Inf) | 0.0008 | 1.00 | 0 | 1f438 |
| great | 🐘 | 5 | 0 | 0 | 0 | 5 | Inf | (2.2,Inf) | 0.006 | 1.00 | 0 | 1f418 |
| great | 🐘👍 | 5 | 0 | 0 | 0 | 5 | Inf | (2.2,Inf) | 0.006 | 1.00 | 0 | 1f44d 1f418 |
| great | 😹 | 4 | 0 | 0 | 0 | 4 | Inf | (1.6,Inf) | 0.02 | 1.00 | 0 | 1f639 |
| great | 💯 | 4 | 0 | 0 | 0 | 4 | Inf | (1.6,Inf) | 0.02 | 1.00 | 0 | 1f4af |
| great | 🌴 | 4 | 0 | 0 | 0 | 4 | Inf | (1.6,Inf) | 0.02 | 1.00 | 0 | 1f334 |
| great | 💪🎉 | 4 | 0 | 0 | 0 | 4 | Inf | (1.6,Inf) | 0.02 | 1.00 | 0 | 1f389 1f4aa |
| great | 💃 | 3 | 0 | 0 | 0 | 3 | Inf | (1.0,Inf) | 0.05 | 1.00 | 0 | 1f483 |
| great | 🏄 | 3 | 0 | 0 | 0 | 3 | Inf | (1.0,Inf) | 0.05 | 1.00 | 0 | 1f3c4 |
| great | 🎈 | 3 | 0 | 0 | 0 | 3 | Inf | (1.0,Inf) | 0.05 | 1.00 | 0 | 1f388 |
| great | 🏳 | 3 | 0 | 0 | 0 | 3 | Inf | (1.0,Inf) | 0.05 | 1.00 | 0 | 1f3f3 |
| great | 🙌👍 | 3 | 0 | 0 | 0 | 3 | Inf | (1.0,Inf) | 0.05 | 1.00 | 0 | 1f44d 1f64c |
| great | 🍑 | 3 | 0 | 0 | 0 | 3 | Inf | (1.0,Inf) | 0.05 | 1.00 | 0 | 1f351 |
| great | 😘🎉🎂 | 3 | 0 | 0 | 0 | 3 | Inf | (1.0,Inf) | 0.05 | 1.00 | 0 | 1f382 1f389 1f618 |
| great | 🏊 | 5 | 1 | 0 | 0 | 6 | 8.9 | (1.3,Inf) | 0.03 | 0.83 | 0.41 | 1f3ca |
| great | 💶 | 5 | 1 | 0 | 0 | 6 | 8.9 | (1.3,Inf) | 0.03 | 0.83 | 0.41 | 1f4b6 |
| great | 👆 | 5 | 1 | 0 | 0 | 6 | 8.9 | (1.3,Inf) | 0.03 | 0.83 | 0.41 | 1f446 |
| great | 🤘 | 13 | 2 | 1 | 0 | 16 | 7.7 | (2.5,Inf) | 0.0003 | 0.78 | 0.48 | 1f918 |
| great | 💡 | 12 | 3 | 0 | 0 | 15 | 7.1 | (2.3,Inf) | 0.0007 | 0.80 | 0.41 | 1f4a1 |
| great | 😚 | 12 | 3 | 0 | 0 | 15 | 7.1 | (2.3,Inf) | 0.0007 | 0.80 | 0.41 | 1f61a |
| great | 💙 | 8 | 2 | 0 | 0 | 10 | 7.1 | (1.7,Inf) | 0.006 | 0.80 | 0.42 | 1f499 |
| great | 💚 | 8 | 1 | 1 | 0 | 10 | 7.1 | (1.7,Inf) | 0.006 | 0.75 | 0.54 | 1f49a |
| great | 🙋 | 13 | 4 | 0 | 0 | 17 | 5.8 | (2.1,Inf) | 0.0008 | 0.76 | 0.44 | 1f64b |
| great | 🍻 | 6 | 2 | 0 | 0 | 8 | 5.3 | (1.2,Inf) | 0.03 | 0.75 | 0.46 | 1f37b |
| great | 👊 | 8 | 3 | 0 | 0 | 11 | 4.7 | (1.4,Inf) | 0.01 | 0.73 | 0.47 | 1f44a |
| great | 💪👍 | 7 | 3 | 0 | 0 | 10 | 4.2 | (1.2,Inf) | 0.03 | 0.70 | 0.48 | 1f44d 1f4aa |
| great | 🙆 | 7 | 2 | 0 | 1 | 10 | 4.2 | (1.2,Inf) | 0.03 | 0.60 | 0.7 | 1f646 |
| great | 🎄 | 11 | 4 | 1 | 0 | 16 | 3.9 | (1.5,Inf) | 0.008 | 0.66 | 0.54 | 1f384 |
| great | 🙃 | 20 | 8 | 2 | 1 | 31 | 3.2 | (1.6,Inf) | 0.001 | 0.58 | 0.61 | 1f643 |
| great | 😝 | 10 | 5 | 0 | 1 | 16 | 3 | (1.1,Inf) | 0.03 | 0.56 | 0.63 | 1f61d |
| great | 🤓 | 13 | 8 | 0 | 0 | 21 | 2.9 | (1.3,Inf) | 0.01 | 0.62 | 0.5 | 1f913 |
| great | 🔝 | 8 | 4 | 1 | 0 | 13 | 2.8 | (1.0,Inf) | 0.05 | 0.58 | 0.57 | 1f51d |
| great | 🙂 | 18 | 11 | 2 | 2 | 33 | 2.1 | (1.1,Inf) | 0.02 | 0.45 | 0.65 | 1f642 |
| great | 🤔 | 15 | 34 | 17 | 13 | 79 | 0.4 | (0.0,0.7) | 0.0009 | -0.08 | 0.64 | 1f914 |
| great | 😢 | 11 | 7 | 28 | 11 | 57 | 0.4 | (0.0,0.8) | 0.005 | -0.25 | 0.68 | 1f622 |
| great | 😭 | 8 | 12 | 24 | 11 | 55 | 0.3 | (0.0,0.6) | 0.0004 | -0.27 | 0.62 | 1f62d |
| great | 😥 | 8 | 11 | 25 | 14 | 58 | 0.3 | (0.0,0.5) | 0.0002 | -0.32 | 0.63 | 1f625 |
| great | 😓 | 6 | 9 | 23 | 5 | 43 | 0.3 | (0.0,0.6) | 0.001 | -0.24 | 0.58 | 1f613 |
| great | 🙄 | 3 | 13 | 4 | 2 | 22 | 0.3 | (0.0,0.8) | 0.02 | -0.05 | 0.53 | 1f644 |
| great | 😔 | 4 | 8 | 16 | 11 | 39 | 0.2 | (0.0,0.5) | 0.0003 | -0.38 | 0.59 | 1f614 |
| great | 😑 | 3 | 3 | 16 | 4 | 26 | 0.2 | (0.0,0.7) | 0.006 | -0.35 | 0.56 | 1f611 |
| great | 😖 | 2 | 4 | 10 | 2 | 18 | 0.2 | (0.0,0.8) | 0.02 | -0.28 | 0.55 | 1f616 |
| great | 🙁 | 2 | 4 | 7 | 5 | 18 | 0.2 | (0.0,0.8) | 0.02 | -0.36 | 0.61 | 1f641 |
| great | 😫 | 2 | 3 | 7 | 7 | 19 | 0.2 | (0.0,0.7) | 0.01 | -0.45 | 0.62 | 1f62b |
| great | 😣 | 2 | 6 | 12 | 7 | 27 | 0.1 | (0.0,0.5) | 0.0008 | -0.41 | 0.54 | 1f623 |

## ACKNOWLEDGMENTS

This work was partially supported by CIT UAE University, the UPAR grant system, Amal Marobuei, Alex Rios and Dani Castro of myHappyForce.com SL who provided the dataset.

## REFERENCES

[1] Fitzpatrick, Kathleen Kara, Alison Darcy, and Molly Vierhile. “Delivering Cognitive Behavior Therapy to Young Adults With Symptoms of Depression and Anxiety Using a Fully Automated Conversational Agent (Woebot): A Randomized Controlled Trial.” JMIR Mental Health 4.2 (2017): e19.

[2] Robinson, Jo, et al. “Social media and suicide prevention: a systematic review.” Early intervention in psychiatry 10.2 (2016): 103-121.

[3] Ljubešic, Nikola, and Darja Fišer. “A global analysis of emoji usage.” ACL 2016 82 (2016).

[4] Martinc, Matej, et al. “Pan 2017: Author profiling-gender and language variety prediction.” Cappellato et al.[13] (2017)

[5] Berengueres, J., & Castro, D. (2017, December). Differences in emoji sentiment perception between readers and writers. In Big Data (Big Data), 2017 IEEE International Conference on (pp. 4321-4328). IEEE.

[6] Kralj Novak, J. Smailovic, B. Sluban, I. Mozetic, Sentiment of Emojis, PLoS ONE 10(12): e0144296, doi:10.1371/journal.pone.0144296, 2015.

[7] Felbo, B., Mislove, A., Søgaard, A., Rahwan, I., & Lehmann, S. (2017). Using millions of emoji occurrences to learn any-domain representations for detecting sentiment, emotion and sarcasm. arXiv preprint arXiv:1708.00524.

[8] Miller, Hannah, et al. "Blissfully happy” or “ready to fight”: Varying Interpretations of Emoji." Proceedings of ICWSM 2016 (2016).Giachanou, Anastasia, and Fabio Crestani.

[9] Hogenboom, A., Bal, D., Frasincar, F., Bal, M., de Jong, F., & Kaymak, U. (2013, March). Exploiting emoticons in sentiment analysis. In Proceedings of the 28th annual ACM symposium on applied computing (pp. 703-710). ACM.

[10] Hirsh, J. B., & Peterson, J. B. (2009). Personality and language use in self-narratives. Journal of research in personality, 43(3), 524-527.

[11] Berengueres, J., Duran, G., & Castro, D. (2017). Happiness an inside Job? Turnover prediction from likeability, engament and relative happiness. IEEE/ACM ASONAM.

[12] Everitt, B. S. (1992). The analysis of contingency tables. Chapman and Hall/CRC.